\documentclass[journal]{IEEEtran}

\usepackage[scaled=1.0]{helvet}
\usepackage{times}
\usepackage{graphicx}
\usepackage{subfigure}
\usepackage{parskip}
\usepackage{multirow}
\usepackage{amsmath}
\usepackage{amssymb}
\usepackage{color}
\usepackage{threeparttable}
\usepackage{multirow}
\usepackage{booktabs}
\usepackage[labelfont=bf,textfont=it]{caption}

\begin{document}
\title{Hybrid Saturation Restoration for LDR Images of HDR Scenes}
%
%
%
\author{
Chaobing Zheng, Zhengguo Li, and Shiqian Wu$^*$
	\thanks{Chaobing Zheng and Shiqian Wu are with the Institute of Robotics and Intelligent Systems,  School of Information Science and Engineering, Wuhan University of Science and Technology, Wuhan 430081, China(e-mails: \{zhengchaobing, shiqian.wu\}@wust.edu.cn).}
	\thanks{Zhengguo Li is with the  SRO Department, Institute for Infocomm Research, Singapore, 138632, (email: ezgli@i2r.a-star.edu.sg).}
}

%
%

\markboth{}
{Shell \MakeLowercase{\textit{et al.}}: Bare Demo of IEEEtran.cls
for Journals}
%



\maketitle

\begin{abstract}
There are shadow and highlight regions in a low dynamic range (LDR) image which is captured from a high dynamic range (HDR) scene.  It is an ill-posed problem to restore the saturated regions of  the LDR image. In this paper, the saturated regions of the LDR image are restored by fusing model-based and data-driven approaches. With such a neural augmentation, two synthetic LDR images are first generated from the underlying LDR image via the model-based approach. One is brighter than the input image to restore the shadow regions and the other is darker than the input image to restore the high-light regions. Both synthetic images are then refined via a novel exposedness aware saturation restoration network (EASRN). Finally, the two synthetic images and the input image are combined together via an HDR synthesis algorithm or a multi-scale exposure fusion algorithm. The proposed algorithm can be embedded in any smart phones or digital cameras to produce an information-enriched LDR image.
\end{abstract}

\begin{IEEEkeywords}
High dynamic range imaging,  saturation restoration, model-based, data-driven, neural augmentation.
\end{IEEEkeywords}

%
\IEEEpeerreviewmaketitle

\section{Introduction}
High dynamic range (HDR) imaging provides realistic viewing experience by capturing scene appearance including lighting, contrasts and details. HDR technique has been already used in a variety of applications on realistic image or video synthesis and thus has been attracted more and more attention \cite{CRFs,Mertens09,1lizg2017}. Most smart phones set the HDR mode as their default mode. Multiple differently exposed low dynamic range (LDR) images are captured to synthesize an HDR image \cite{CRFs}. When such a set is captured by a handhold device with a conventional sensor, there might be cameral movement or moving objects in the captured images. Even though many alignment and ghost removal techniques have been proposed \cite{1wu2014,Hu12,Sen12,Li,1kala2017,1wu2018,1lih2020}, both topics, especially ghost removal are still issues. The ghost artifacts  can be solved by specialized high-end imaging devices \cite{1tocci2011,1Gu2010}.

One alternative solution is proposed in this paper to capture one LDR image by using a smart phone or a digital camera and restore the saturated regions of the LDR image to generate an information-enriched LDR image. Generally, there are shadow and highlight regions in the LDR image.  It is desirable to restore both the shadow and the highlight regions for the LDR image. To this end, two synthetic images, one with a smaller exposure, the other with a larger exposure are generated to restore the highlight and shadow regions, respectively. Clearly, the proposed method is an indirect method which is different from existing direct methods in \cite{ExpandNet,SingleHDR,1santos2020}. The proposed indirect method allows a user to combine all the three images into an HDR image or an LDR image according to her/his preference. It is noted that the data-driven reverse tone mapping algorithms in \cite{DrTMO,1kim2021} are also on top of the indirect method.  In this paper, the saturation is restored by fusing data-driven and model-based approaches rather than the data-driven approaches in \cite{DrTMO,1kim2021}. As indicated in \cite{1nir2021}, such a neural augmentation must posses the complete domain knowledge it requires as in \cite{1zheng2020}. The camera response functions (CRFs) are thus assumed to be known in advance. In other words, the proposed algorithm is supposed to be embedded into a smart phone or a digital camera as the usage in \cite{1chen2018}.

The synthetic images are first generated by a model-based approach and then refined by using a data-driven approach. The combination of the model-driven and data-driven methods forms a neural augmentation \cite{1nir2021}. The synthetic image are initially generated by using intensity mapping functions (IMFs) to avoid  lightness distortion  \cite{LECARM} if the IMFs are reliable \cite{Li}. Otherwise,  the fixed ratio strategy in \cite{1lizg2017} is applied to reduce the color distortion. The generation of the initial dark image is focused on because the generation of the initial brighter image has been studied in \cite{1zheng2020}.  The initially synthetic images are refined by a  novel exposedness aware saturation restoration network (EASRN) which is proposed by incorporating an exposedness aware guidance branch (EAGB) into a nonlocal recursive residual group (NRRG). As pointed out in \cite{1zheng2020}, the objective of \cite{1zheng2020} is to explore the feasibility of the neural augmentation framework rather than a sophisticated deep neural network (DNN) for single image brightening. The DNN in \cite{1zheng2020} is a ResNet \cite{1he2016} with multiple residual blocks. Intrinsic properties of differently exposed images are ignored by the DNN in  \cite{1zheng2020} while they are well utilized by the proposed EASRN. The exposedness of the input image is used to define two binary masks. The proposed masks are based on the intrinsic properties of differently exposed images: \cite{1zheng2013,Li}: 1) the IMF from the input image to the dark image is not reliable for all pixels in the overexposed regions of the input image; and 2) the IMF from the input image to the bright image is not reliable for all pixels in the underexposed regions of the input image.  Clearly, they are different from the soft mask in \cite{1santos2020}. The properties are utilized to design the EAGB.  The objective of the EAGB is to help the proposed EASREN pay more attention to the saturated regions in order to restore them efficiently.  The NRRG is inspired by the nonlocal block in \cite{1wang2018, 1fu2019} and the recursive residual group module in \cite{CycleISP} which explores both local and non-local correlation along both spatial dimension and channel dimension of the synthetic images. As shown in \cite{1buda2005,1dabov2007}, each block in the input image could have nonlocal similar blocks. The self-similarity of the input image could be applied to refine the shadow and highlight regions in the initially synthetic images. The proposed algorithm restores the shadow and highlight regions from both the neighboring areas and the nonlocal similar regions while the algorithm in \cite{1zheng2020} restores the shadow regions only from the neighboring areas. The two synthetic images and the input image are combined together via the state-of-the-art multi-scale exposure fusion (MEF) algorithm  in \cite{TOMP} or the HDR synthesis algorithm in \cite{CRFs}. Experimental results indicate that the EASRN outperforms the DNN without the EAGB. The proposed neural augmentation also outperforms the hybrid learning method in \cite{1zheng2020}. This implies that the intrinsic properties of differently exposed images are indeed useful for the DNN. The proposed algorithm can produce better LDR images than the state-of-the-art algorithms in terms of subjective and objective judgements. In summary, the contributions of this paper are highlighted as follows:

1) Our work is the first to restore both underexposed and overexposed regions of the LDR image by fusing model-based and data-driven approaches. The model-based and data-driven saturation restoration {\it compensates} each other, and form a neural augmentation. Such a neural augmentation has a potential to achieve learning with few data \cite{1nir2021}.

2) A novel EASRN is proposed by incorporating  the EACB into the NRNG to improve its performance. The intrinsic properties of differently exposed images are well utilized by the EASRN.

The rest of this paper is organized as below. Relevant works on HDR imaging  are reviewed in Section \ref{new}. The proposed saturation restoration algorithm is presented in Section \ref{paradigm1}.  Extensive experimental results are provided in Section \ref{paradigm4} to verify the proposed algorithm. Finally, conclusion remarks are drawn in Section \ref{paradigm5}.

\section{Relevant Works on HDR Imaging}
\label{new}
An image captured by an ordinary smart phone or digital camera is typically represented by eight-bit integers which are far from adequate
as the visible radiance in a real high dynamic range (HDR) scene has a dynamic range larger than $10^4$. An efficient solution to address the problem
is to capture a set of differently exposed images \cite{CRFs} and to integrate them together to generate an image that contains
the rich information from all of the input images. There are two kinds of solutions to generate a content rich image. One is to synthesize multi-exposed images into an HDR image \cite{CRFs}, and then compress the HDR image into an LDR image \cite{1reinhard2002,1durand2002,1farbman2008,1liang2018} such that it can be displayed by conventional LDR digital devices. The other is to fuse all of the differently exposed LDR images into a single LDR image \cite{Mertens09,1lizg2017,Ancuti2017,TOMP,Y,YangY2020,MEFNet}. The latter dominates the HDR solution in smart phones.

Both types of methods require the captured scene to be static and the camera to be stationary. The movement of camera would cause the fused image to be blurred while any movement of objects would contribute to a phenomenon called ghosting artifact in the final synthesized output. It is relative easy to align differently exposed images, for example, by the method in \cite{1wu2014}. It is much more challenging to prevent ghosting artifacts from appearing in the final image, and ghost artifacts are thus believed to be Achilles's heel for the HDR imaging. The problem could be addressed by using special capturing systems.  One example is a beam splitting based HDR  capturing system with few sensors \cite{1tocci2011}.  Another one is a row-wise CMOS HDR video capturing system \cite{1Gu2010,1xu2021}. Canon also released an innovative global shutter with a specific sensor that reads the sensor twice in an HDR mode \cite{1cannon2019}.

Besides the new capturing systems, one more attractive solution is to capture one LDR image by using a smart phone or a digital camera and restore an HDR image from the LDR image. This solution is known as reverse tone mapping \cite{1bant2006} and the reverse tone mapping is an ill-posed problem. A simple linear expansion method was proposed in \cite{1akyuz2007}. Psychophysical experiments in \cite{1akyuz2007} show that such simple operation has a potential to outperform a true HDR image. Recently, an indirect approach was proposed in \cite{DrTMO}  to infer a sequence of synthetic LDR images with different exposures via a deep learning based method, and then reconstruct a pseudo-HDR image by merging the LDR images using the algorithm in \cite{CRFs}. Image formation pipeline  including the dequantization, linearization, and hallucination was reversed in \cite{SingleHDR} to track single-image HDR reconstruction problem. The difficulty of training one single network for reconstructing HDR images is significantly reduced. A data-driven approach was proposed in \cite{1santos2020}  to reconstruct an HDR image by recovering the saturated pixels of an input LDR image, and a soft feature masking mechanism was introduced to reduce artifacts caused by the invalid information in the saturated regions.  The pseudo-HDR image needs to be compressed by using the existing tone mapping algorithms \cite{1reinhard2002} such that it can be displayed by most existing digital devices. The complexity of the pseudo-HDR image could be an issue. It is thus desire to develop a simpler algorithm to restore an LDR image of an HDR scene from the saturation point of view.

Single image brightening also intends to produce an information-enriched LDR image. The input is an image captured by a small exposure time and a small ISO value \cite{1lizg2017}. The single image brightening algorithms can be classified into three categories. One category is model-based methods, such as low lighting image enhancement (LIME) \cite{LIME}, and algorithms in \cite{1lizg2017,LECARM}. Another category is deep learning based ones \cite{1chen2018,DeepUPE}.  The other is neural augmented methods which fuse both model-based and data-driven methods \cite{1zheng2020}. The difference between single image brightening and saturation restoration is that only the shadow regions are restored by the single image brightening while both the shadow and the highlight regions are restored by the saturation restoration. Noise amplification is a challenge for the former while both noise amplification and color distortion are challenges for the latter. Our human eyes are more sensitive to the color distortion in the bright regions.
Intrinsic properties of differently exposed images are ignored by the DNN in  \cite{1zheng2020} but they will be well utilized by the proposed framework.

\begin{figure*}[htb]
	\centering
	\includegraphics[width=0.88\textwidth]{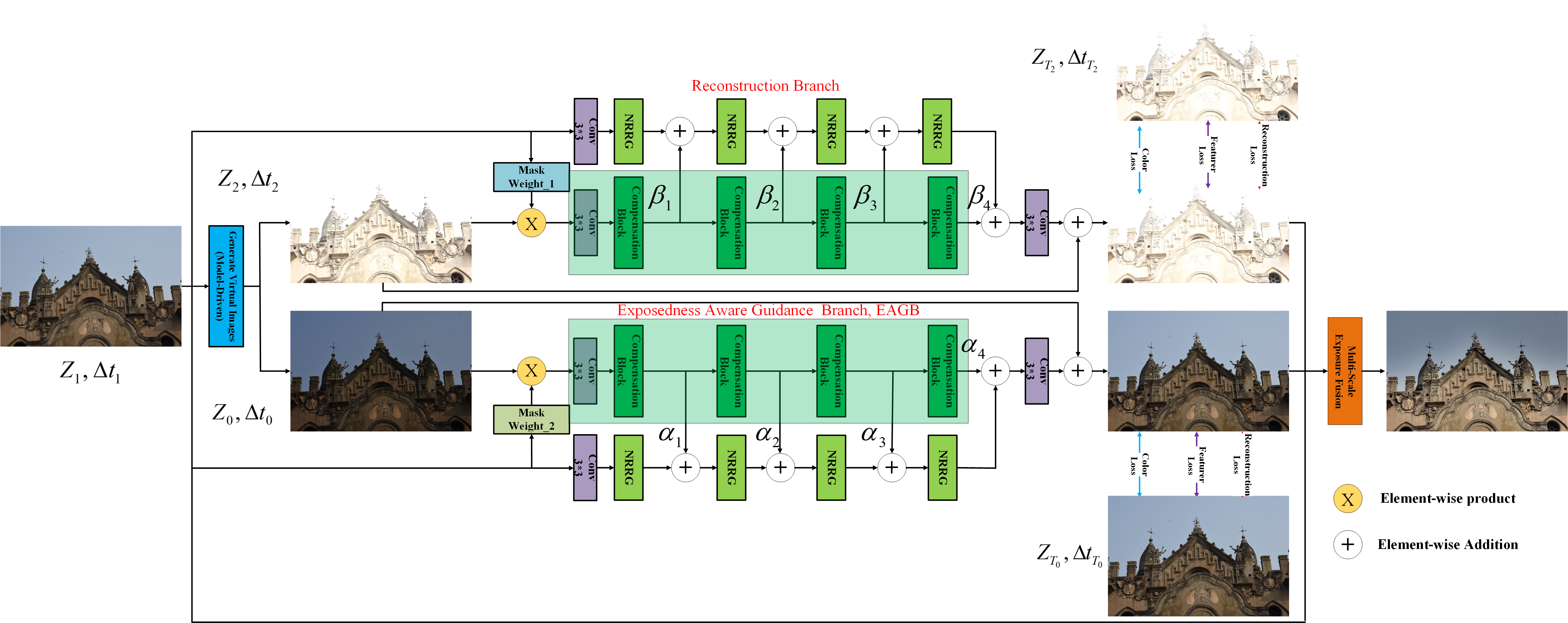}
	\caption{The diagram of the proposed saturation restoration algorithm via fusion of model-based and data-driven approaches. Two synthetic images $Z_0$ and $Z_2$ with ${\Delta} t_0$ and ${\Delta} t_2$ are first generated by a model-based method, i.e. the intensity mapping function (IMF) based method. $({Z_{T_0}}-{Z_0})$ and $({Z_{T_2}}-{Z_2})$ are then learnt via the proposed EASRN to enhance the initial synthetic images. The input image and the two synthetic images are finally fused to obtain an information enriched LDR image.}
	\label{Fig_1}
\end{figure*}

\begin{figure*}[htb]
	\centering
	\includegraphics[width=0.88\textwidth]{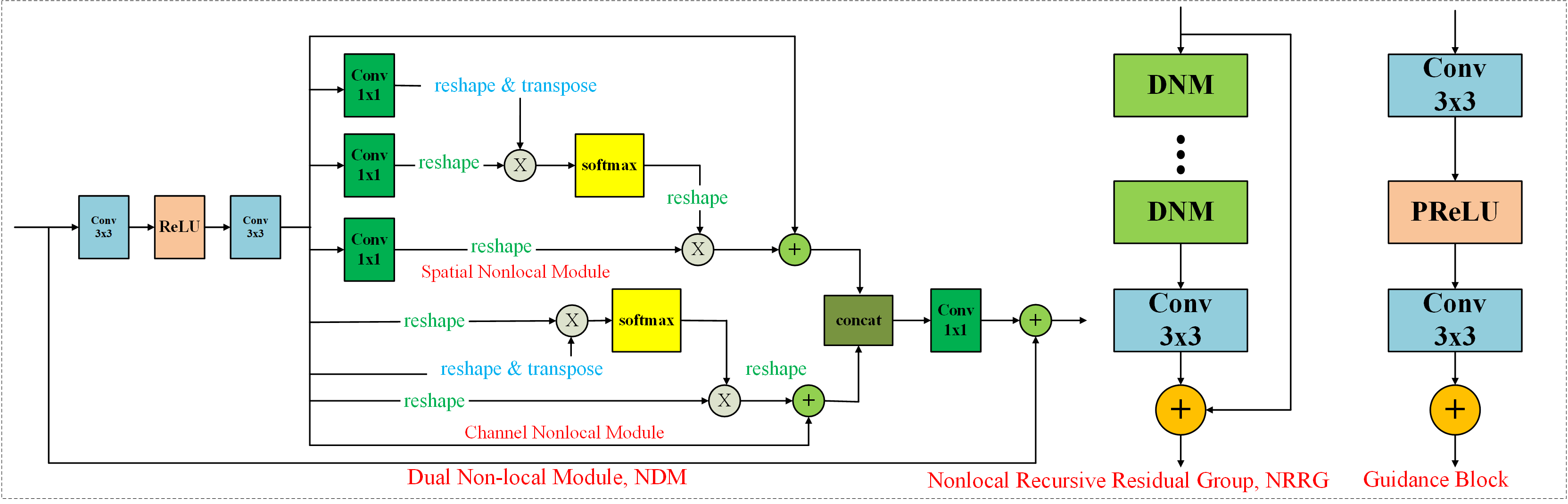}
	\caption{Nonlocal recursive residual group (NRRG) contains multiple non-local dual  modules (NDMs). Each NDM contains non-local spatial and channel  modules. The guidance block includes two $3*3$ convolution layers and one active layer.}
	\label{Fig1B}
\end{figure*}

\section{Neural Augmented Saturation Restoration}
\label{paradigm1}

Let $Z_1$ be an eight-bit image.  The key component of the proposed algorithm is to generate two high-quality 8-bit synthetic images $Z_0$ and $Z_2$ by fusing a model-based method and a data-driven one. The image $Z_0$ is darker  than the input $Z_1$ to restore the highlight regions of $Z_1$ while the image $Z_2$ is brighter than $Z_1$ to restore the shadow regions of $Z_1$.  The proposed framework is a neural augmentation. As indicated in \cite{1nir2021}, such a framework must posses the complete domain knowledge it requires. Thus, the CRFs are assumed to be known as in \cite{1chen2018,1zheng2020}. In other words, each algorithm is trained for each camera.

 The ground truth images of $Z_0$ and $Z_2$ are denoted as $Z_{T_0}$ and $Z_{T_2}$, respectively. They are captured together with the image $Z_1$ by using the method in \cite{CRFs}.  Fig. \ref{Fig_1} summarizes the pipeline of the proposed saturation restoration algorithm. The synthetic images are first generated by using the IMFs. They are then refined by the proposed EASRN which is on top of the RRG in \cite{CycleISP} and the nonlocal block in \cite{1wang2018}. Finally, the input image $Z_1$ and two synthetic images $Z_0$ and $Z_2$ are fused together using the MEF algorithm in \cite{TOMP} or the HDR synthesis algorithm in \cite{CRFs} to produce the final image.

\subsection{Model-Based  Saturation Restoration}
\label{paradigm2}

The two synthetic images  $Z_0$ and $Z_2$ are first produced by using the IMFs.  Since the details on generating the image $Z_2$ via the IMFs are available in \cite{1zheng2020}, this subsection focuses on the generate of the image $Z_0$ via the IMF.

Let the CRF be ${f_l}(\cdot)(l\in\{R, G, B\})$, and the exposure times of  $Z_0$ and $Z_1$ and $Z_2$ be ${{\Delta}t_0}$, ${{\Delta}t_1}$ and ${{\Delta}t_2}$, respectively. Without loss of generality, it is assumed that  ${{\Delta}t_2} > {{\Delta}t_1} > {{\Delta}t_0}$. It has been shown in \cite{Y} that there is relative brightness change in the fused image if the exposure ratios are too large. Thus, $\Delta t_2$ and $\Delta t_0$ are selected as $4\Delta t_1$ and $\Delta t_1/4$ so as to restore the information as much as possible. The IMF between the input image $Z_1$ and the synthetic image $Z_0$ can be expressed as follows:
\begin{equation}
\label{IMFS}
{\Lambda}_{1\to 0,l}(z)={f_l}(\frac{{f^{-1}_l}(z){{\Delta}t_0}}{{\Delta}t_1}),
\end{equation}
where  ${f^{-1}_l}(\cdot)$ is the inverse function of the CRF ${f_l}(\cdot)$.

Instead of using the fixed ratio strategy in \cite{1lizg2017}, the initial synthetic image $Z_0$ is generated by using the IMFs if they are reliable. The lightness distortion is prevented from appearing in the darkened images. As mentioned in \cite{1zheng2013,Li}, if the pixel value $z$ is smaller than a threshold ${\xi}_U$,  ${\Lambda}_{1\to 0,l}(z) $ is a one-to-one mapping, which is reliable. Otherwise, it is not reliable due to a one-to-many mapping. Both cases are considered to produce the two initial synthetic images as follows:

{\it Case 1}: All ${Z_{1,l}(p)}$'s are smaller than the threshold ${\xi}_U$ for all the three channels. The pixel values corresponding to the synthetic image $Z_0$ can be computed by using the IMF as
\begin{align}
Z_{0,l}(p)&=\Lambda_{1\to 0,l}({Z_{1,l}(p)})\; ;\; l\in\{R, G, B\}.
\end{align}

{\it Case 2}: ${Z_{1,l}(p)} $ is larger than the  threshold ${\xi}_U$ for at least one channel, and the IMF is not reliable, which yields color distortion. Hence, the fixed ratio strategy in \cite{1lizg2017} is adopted to generate the corresponding synthetic pixels. One challenging problem to be addressed is that there are visible seams at the boundaries among the highlighted  regions and their neighboring regions if the ratio is selected as in \cite{1lizg2017}.  To address the second problem, the synthetic pixel $Z_0(p)$ is computed as
\begin{equation}
\label{decomposed}
Z_0(p)= \tilde{\gamma}_0  Z_{1}(p),
\end{equation}
where the value of  ${{\tilde{\gamma}}_0}$ is obtained by minimizing the following function:
\begin{equation}
\label{mini}
\sum_{l\in \{R, G, B\}}{\tilde{w}(Z_{1,l}(p))(\Lambda_{1\to 0,l}(Z_{1,l}(p))-{{\tilde{\gamma}}_0}{Z_{1,l}(p)})^2},
\end{equation}
and the function $\tilde{w}$ is defined as \cite{1yao2012}:
\begin{eqnarray}
&&\hspace{-7mm}\tilde{w}(z) = \left\{ {\begin{array}{*{20}{l}}
	{0;}&{if~}{{\rm{255}} { \geq z > {\xi_U}}}\\
	{128 - 3{h_1}^2(z) + 2{h_1}^3(z);}&{if~}{{\xi_U} \geq z > {\xi_L}}\\
	{128;}&{otherwise}{}.
	\end{array}} \right.,
\end{eqnarray}
The function ${h_1}(z)$ is given as
\begin{equation}
\label{h}
{h_1}(z)=\frac{\xi _U-z}{\xi _U-\xi _L}.
\end{equation}
Similar to $\xi_U$, the value of $\xi_L$ depends on camera quality. The higher the quality of the camera, the larger the value of $\xi _U$. In this study, we use $\xi _U=250$ and $\xi _L=200$  as the default settings.

\begin{figure*}[htb]
	\centering
	\includegraphics[width=0.88\textwidth]{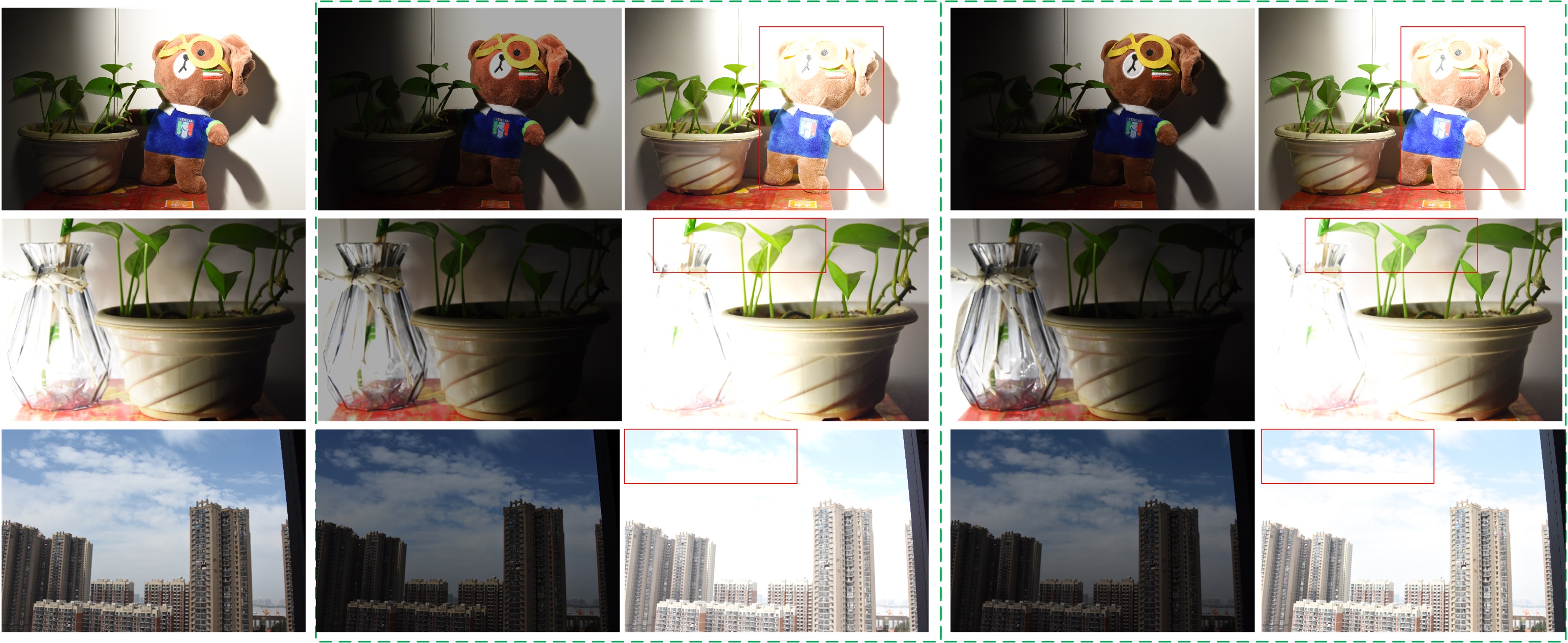}
	\caption{The first column includes three input LDR images which are taken with a Nikon 7200 camera. The ISO value is set as  800, and the exposure time ${\Delta} t_1$ is very short. The second column is the set of initial synthetic images $Z_0$'s, the exposure time is ${\Delta} t_0$.  The third column is the set of initial synthetic images $Z_2$'s, whose exposure time is ${\Delta} t_2$,  ${\Delta} t_0 < {\Delta} t_1 < {\Delta} t_2$. The fourth column is the set of ground truth images $Z_{T_0}$ with the exposure time  ${\Delta} t_0$. The fifth  column is the set of ground truth images $Z_{T_2}$ with the exposure time ${\Delta} t_2$.  }
	\label{Fig3}
\end{figure*}

Since the highlighted region of $Z_1$ is not as noisy as the shadow region of $Z_1$, the pixel $Z_1(p)$ in the highlighted region is not decomposed into two layers as in \cite{1zheng2020}. The resultant synthetic images are shown in Fig. \ref{Fig3}. The synthetic images are close to the corresponding ground truth images if the IMFs are reliable. However, due to limited representation capability of IMFs, the images $Z_0$ and $Z_2$ do not contain all the information in the ground truth images, especially for those saturated regions. Both intensity and color need to be improved for the underexposed regions of the image $Z_2$ and the overexposed regions of the image $Z_0$.

\subsection{Data-Driven Saturation Restoration}
\label{paradigm3}

The IMFs can be regarded as a model to represent the correlation among different exposed images but its representation capability is limited. The unmodelled information $({Z_{T_0}}-{Z_0})$ can be further represented by a data-driven method. The unmodelled information is usually sparse, i.e.,  most values are likely to be zero or small. It can be easily {\it compensated} by the data-driven method. Subsequently, the saturated regions are further restored. It is worth noting that the modelled information and un-modelled information are analogous to the modelled dynamics and un-modelled dynamics in the field of control systems \cite{1khalil2002}.

An EASRN is proposed to refine the images $Z_0$ and $Z_2$ both locally and non-locally. As shown in Figs. \ref{Fig_1} and \ref{Fig1B}, the proposed EASRN consists of EAGB and NRRG. The EAGB can help the NRRG pay more attention to saturated areas, which can guide the saturation restoration more effectively. Two binary masks $M_{0,l}$ and $M_{2,l}$ are defined for the image $Z_{1,l}$ as
\begin{align}
\label{M0}
&M_{0,l}(p) = \left\{\begin{array}{ll}
	0;&\mbox{if~} Z_{1,l}(p) \geq \xi_U\\
	1;&\mbox{otherwise}
	\end{array} \right.,\\\label{M2}
&M_{2,l}(p) = \left\{ \begin{array}{ll}
	0;&\mbox{if~} Z_{1,l}(p) \leq \xi_L\\
	1;&\mbox{otherwise}
	\end{array} \right..
\end{align}

The different masks $M_{0,l}$ and $M_{2,l}$ will be used to refine the initial images $Z_0$ and $Z_2$, respectively. They are defined  by using the following intrinsic properties of differently exposed images \cite{1zheng2013,Li}:

1) The IMF $\Lambda_{1\to 0,l}(\cdot)$  is not reliable for all pixels in the overexposed regions of the image $Z_{1,l}$; and 2) The IMF $\Lambda_{1\to 2,l}(\cdot)$ is not reliable for all pixels in the underexposed regions of the image $Z_{1,l}$.

Obviously, the masks $M$'s are derived from intrinsic properties of differently exposed images, and they are different from the soft mask in \cite{1santos2020}.

The masks (\ref{M0}) and (\ref{M2}) are utilized to design the EAGB.  They are first multiplied with the synthetic images to filter out the poor information from the ill-posed regions. The features of the EAGB are embedded into the NRRG in a multilevel manner to realize deep guidance. Subsequently, the proposed EASRN can pay necessary attention to the saturated areas for the saturation restoration. Clearly, the interpretability of the EASRN is improved with the proposed EAGB.

The function of the NRRG is to globally suppress the less useful features and only allow the propagation of more informative ones. The NRRG contains multiple non-local dual  modules (NDMs) \cite{1wang2018,1fu2019} and each NDM performs both  non-local  spatial and channel operations. Let $X\in \mathbb{R}^{C\times H\times W}$ be the input feature map. The details on the spatial nonlocal module and channel nonlocal module are given as below.

{\bf Non-local spatial  module}: The feature map $X$ is fed into a convolution layers to generate three new feature maps $\hat{X}^s_1(\in \mathbb{R}^{C\times H\times W})$, $\hat{X}^s_2(\in \mathbb{R}^{C\times H\times W})$ and $\hat{X}^s_3(\in \mathbb{R}^{C\times H\times W})$, respectively. Let $N=H\times W$. All of them are shaped into $\mathbb{R}^{C\times N}$. A matrix multiplication between the transpose of $\hat{X}^s_1$
and $\hat{X}^s_2$, and  a softmax layer are then applied to calculate the spatial similarity map $S^s\in \mathbb{R}^{N\times N}$:
\begin{align}
S^s_{ji}=\frac{\exp(\hat{X}^{s}_{1,i}\cdot \hat{X}^{s}_{2,j})}{\sum_{i=1}^N \exp(\hat{X}^{s}_{1,i}\cdot \hat{X}^{s}_{2,j})}.
\end{align}
The non-local spatial module is defined as follows:

\begin{align}
E^s_j=\sum_{i=1}^N ({S^s}_{j,i}\cdot \hat{X}^{s}_{3,i}) + X_j
\end{align}

{\bf Non-local channel  module}: The feature map $X$ is shaped into $\mathbb{R}^{C\times N}$. A matrix multiplication between $X$ and the transpose of $X$, and a softmax layer are then applied to calculate the channel similarity map $S^c\in \mathbb{R}^{C\times C}$:
\begin{align}
S^c_{ji}=\frac{\exp(X_i\cdot X_j)}{\sum_{i=1}^C \exp(X_i\cdot X_j)}.
\end{align}
Then, the non-local channel module is defined as follows:
\begin{align}
E^c_j=\sum_{i=1}^C (S^c_{j,i}\cdot X_i) + X_j.
\end{align}

Besides the structure of the proposed network, loss functions also play an important role in the data-driven approach. The proposed loss function is defined as
\begin{equation}
L= L_r+w_cL_c + w_fL_f,
\end{equation}
where $w_c$ and $w_f$ are constant, they are selected as $0.01$ and $0.01$ in this study, respectively.

The restoration loss $L_r$ is usually defined as
\begin{equation}
L_r = \sum_{p,l}[Z_{T_i,l}(p)-Z_{i,l}(p)-{f_i}(Z_{1,l}(p))]^2.
\end{equation}
Since the under-exposed regions in $Z_1$ contain random noise \cite{1lizg2017}, a content adaptive weight is introduced to the restoration loss so as to reduce the effect of noise on the adjustment parameters. The loss function $L_r$ is given as
\begin{equation}
\label{eq15}
L_r = \sum_{p,l}W_{i,l}(p) [Z_{T_i,l}(p)-{f_i}(Z_{1,l}(p))-Z_{i,l}(p)]^2,
\end{equation}
where the weight function $W_{i,l}(p)$ is expressed as:
\begin{eqnarray}
&&\hspace{-7mm}{W_{i,l}}(z) = \left\{ {\begin{array}{*{20}{l}}
	{1;}&{if~} Z_{i,l}(p)\geq \nu\\
	{\frac{1}{\nu-{Z_{i,l}(p)}};}&{otherwise}{}.
	\end{array}} \right.,
\end{eqnarray}
and $\nu$ is a small positive constant and it is empirically selected as 6.0 in this paper if not specified. When the pixel value in position $p$ is smaller than $v$, it may be noise, so a small weight is assigned to the loss.

To minimize the possible color distortion in the two synthetic images, the color loss is defined as
\begin{equation}
\label{a1}
L_c=\sum_p\angle (Z_{T_i}(p),  Z_i(p)+{f_i}(Z_{1}(p))),
\end{equation}
where $\angle(Z_{T_i}(p),{f_i}(Z_1(p))+Z_i(p))$ is the angle between two 3D $(R, G, B)$ vectors $Z_{T_i}(p)$ and $ ({f_i}(Z_1(p))+Z_i(p))$. Since the $L_r$ metric only measures the color difference numerically, it cannot ensure that the color vectors have the same direction \cite{DeepUPE}. By employing the color loss in Eq. (\ref{a1}), the possible color distortion is reduced, especially for the saturate regions.

The feature-wise loss $L_f$ is defined as
\begin{align}
L_f=\frac{1}{W_{i,j}}\frac{1}{H_{i,j}}\sum_{l=1}^{W_{i,j}}\sum_{m=1}^{H_{i,j}}(\phi_{i,j}(y)_{l,m}-\phi_{i,j}(y_0+\tilde{f}(y_0))_{l,m})^2,
\end{align}
where $W_{i,j}$ and $H_{i,j}$ denote  the dimensions of the respective feature maps within the VGG network. $\phi_{i,j}(\cdot)$ is the feature map obtained by the $j$-th convolution (before activation) before the $i$-th maxpooling layer within the VGG network. Instead of using commonly adopted feature-wise loss function that adopts a VGG network trained for image classification, a fine-tuned VGG network for material recognition in \cite{1bell2015} is adopted to define the feature-wise loss. The VGG in \cite{1bell2015} focuses on textures rather than object and the texture is critical for the refinement of the synthetic images.

The enhanced synthetic images is expressed as $({f_i}(Z_1)+Z_i)( i=0,2)$. They are much closer to the ground truth images, which implies the unmodelled information is reduced significantly. The proposed algorithm has the following advantages: Firstly, compared with the model-based method, the synthetic images produced by the proposed method is enhanced by compensating unmodelled information. Secondly, compared with the data-driven solution, the proposed framework converges fast and requires fewer training samples, because $({Z_{T_0}}-{Z_0})$ and $({Z_{T_2}}-{Z_2})$ are sparser than ${Z_0}$ and ${Z_2}$. Thirdly, the EAGB can assist the network to pay considerable attention to the well-exposed areas, which can provide accurate information. Thus, it is easy to train the proposed EASRN. Clearly, the model-based and data-driven methods can {\it compensate} each other.

\subsection{Combination of Input Image and Two Synthetic Images}
\label{fusion}
The input image $Z_1$ and the two synthetic images $Z_0$ and $Z_2$ can be combined together by using the HDR synthesis algorithm in \cite{CRFs} to produce an HDR image or by using the MEF algorithm in \cite{TOMP} to produce a LDR image. The latter is focused on in this subsection.

As shown in  \cite{TOMP}, the weighting maps of the differently exposed images play an important role in the MEF algorithm. This subsection focuses on defining the weighting maps for the three images. One function is used to determine amplification factors of all the pixels in the image $Z_1$. The weight is defined as
\begin{align}
\psi_1(z)=2.
\end{align}

The other is to measure contrast, well exposedness level, and color saturation of each pixel in the three images and it is defined the same as in \cite{TOMP}
\begin{align}
\psi_2(Z_i(p)) = w_c(Z_i(p))\times w_s(Z_i(p))\times w_e(Z_i(p)),
\end{align}
where $w_c$ is computed by filtering the gray-scale version of each image via a Laplacian filter. $w_s$  is the standard deviation among the
R, G and B channels. $w_s$ is derived by first using a Gauss curve on each color channel separately and then multiplying the results.

Let $Y_1$ be the luminance component of the image $Z_1$ in YUV color space. The weight of the pixel $Z_1(p)$ is given as
\begin{align}
\label{weight1}
W(Z_1(p)) = \psi_1(Y_1(p))\psi_2(Z_1(p)),
\end{align}
and the weight of the pixel $Z_i(p)(i = 0,2)$ is given as
\begin{align}
\label{weight2}
W(Z_i(p)) = \psi_2(Z_i(p)).
\end{align}

With the above weighting maps, all the images $Z_i(i=0,1,2)$ are fused together via the state-of-the-art MEF algorithm  in \cite{TOMP}. The number of pyramid levels is increased by one to reduce the halo artifacts. The MEF-SSIM is also slightly improved due to the small size of the images.

\section{Experimental Results}
\label{paradigm4}
Extensive experiments are carried out in this section to demonstrate the rationality and effectiveness of the proposed neural augmentation framework. The emphasis is to show how the model-based method and the data-driven one {\it compensate} each other. Readers are invited to view to the electronic version of the full-size figures and zoom in these figures in order to better appreciate the differences among images.


\begin{figure*}[t]
	\begin{center}
		\includegraphics[width=0.88\linewidth]{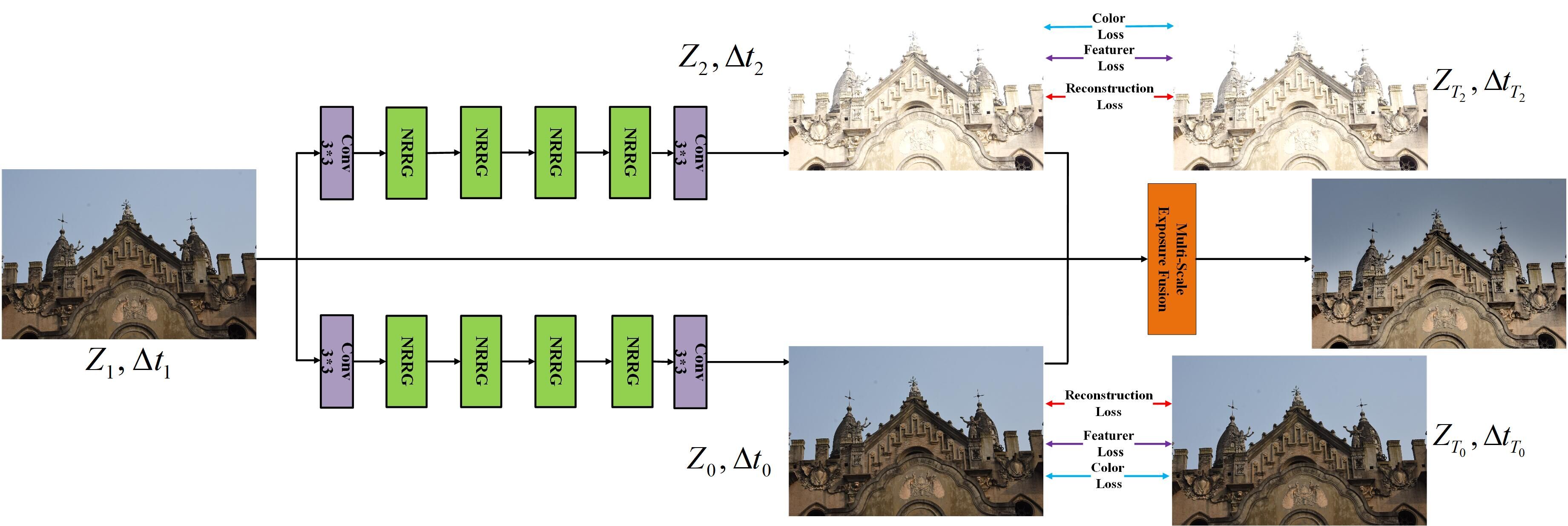}
	\end{center}
	\caption{The diagram of a data-driven  saturation restoration algorithm. Two synthetic images $Z_0$ and $Z_2$ with ${\Delta} t_0$ and ${\Delta} t_2$ are generated directly by data-driven method. The input image and the two synthetic images are finally fused to obtain an information enriched LDR image. }
	\label{Data_Driven_Frame}
\end{figure*}

\subsection{Dataset}

We have built up a dataset which comprises 420 sets of differently exposed image. Each set contains three images. The images from difference scenes are captured using a Nikon 7200 camera. The ISO is set as  800.  According to \cite{CRFs}, exposure times are different while other configurations of the camera are fixed. The interval of exposure ratio between them is 2 exposure values (EVs). Both camera shaking and object movement are strictly controlled to ensure that only the exposure time is different.  All the images in the dataset are randomly split into two parts: 340 images sequences for training and the rest 80 sequences for testing.

The number of NRRGs and guidance blocks are set as $4$, and each NRRG contains $4$ NDMs. We randomly crop a $128*128$ patch from each input during training. The proposed network is trained using the proposed loss functions and Adam optimizer with ${{\beta}_1}=0.9$ and ${{\beta}_2}=0.99$. We set the batch size to $8$. The learning rate is initially set to $10^{-4}$ and then decreased using a cosine annealing schedule. All the experiments are implemented using PyTorch on NVIDIA GP100 GPUs.

\subsection{Evaluation of The EAGB}
\begin{figure}[htb]
	\centering
	\subfigure{
		\begin{minipage}[b]{0.465\linewidth}
			\includegraphics[width=1\linewidth]{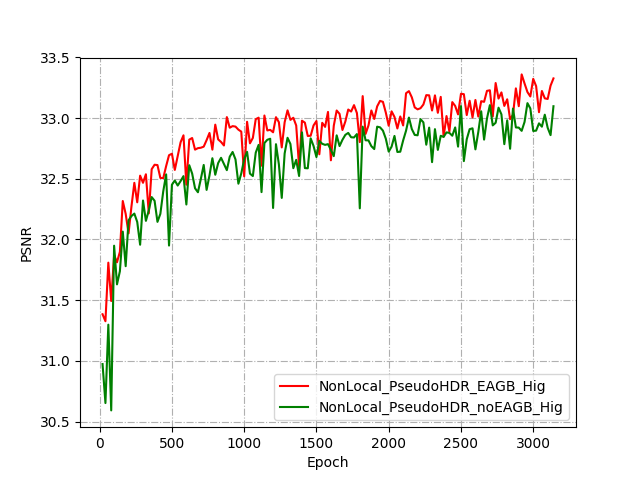}	
			\centerline{(a)}	
	\end{minipage}}
	\subfigure{
		\begin{minipage}[b]{0.465\linewidth}
			\includegraphics[width=1\linewidth]{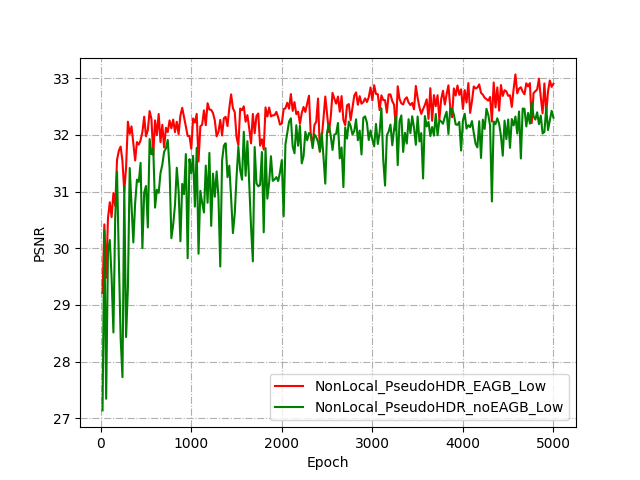}
			\centerline{(b)}
	\end{minipage}}
	\caption{(a) Comparisons of PSNR between the modes with and without the EAGB for high exposure images. (b) Comparisons of PSNR between the modes with and without the EAGB for low exposure images. The mode with the EAGB is more stable than the mode without EAGB, and can also achieve higher PSNR. }
	\label{EAGB}
\end{figure}

 The features from the EAGB are embedded into the NRRG to guide the NRRG.  In order to validate the effectiveness of the EAGB,  the two modes with and without the EAGB are compared. As shown in Fig. \ref{EAGB}, the mode with the EAGB is more stable than the mode without EAGB, and can achieve higher PSNR values in different epoches. Clearly, the EAGB indeed guides the NRRG to exploit  reliable information for restoring the synthetic images. The intrinsic properties of differently exposed images are indeed helpful for the data-driven approach.

The values of ${\alpha}_i$ and ${\beta}_i$ are shown in Table \ref{tab1}. The values of ${\alpha}_1$ and ${\beta}_1$ are $1.0112$ and $0.9884$, others are approximately $1.0$ which means that the proposed EAGB provides useful information for the saturation restoration of LDR images. The values of ${\alpha}_4$ and ${\beta}_4$ are bigger than $1.0$.  The extracted features are still important for the saturation restoration when the network goes deeper.

\begin{table}[htb]
	\begin{center}
		\centering
		\caption{$\alpha$ and $\beta$ in the trained EAGB}
		\tabcolsep8pt\begin{tabular}{ccccccccccccc}
			\hline		
			\multirow{1}*{${\alpha}_1$}   &  ${\alpha}_2$    &  ${\alpha}_3$  &   ${\alpha}_4$  \\
			\multirow{1}*{ 0.9884} & 0.9849   & 0.981   & 1.0826 \\
			\hline
			\multirow{1}*{${\beta}_1$}    &  ${\beta}_2$     &  ${\beta}_3$   &   ${\beta}_4$   \\
			\multirow{1}*{1.0112}  & 0.9789   & 0.9888   & 1.0021  \\
			\hline
		\end{tabular}
		\label{tab1}
	\end{center}
\end{table}

\begin{table}[htb]
	\centering
	\begin{threeparttable}
		\caption{SSIM and PSNR of synthetic images generated by different methods.}
		\begin{tabular}{ccccccc}
			\toprule
			\multirow{2}{*}{}&
			\multicolumn{2}{c}{ SSIM }&\multicolumn{2}{c}{ PSNR }\cr
			\cmidrule(lr){2-3} \cmidrule(lr){4-5}
			& Low  &  Hig &  Low & Hig\cr
			\midrule
			model-based                                        &0.9442  &0.94   &26.67   &29.63    \cr
			Hybrid learning in \cite{1zheng2020}               &0.9469  &0.9477   &32.40   &32.60    \cr
			data-driven in Fig. \ref{Data_Driven_Frame}        &0.9531  &0.9507   &32.55   &32.77     \cr
			proposed (no EAGB)                                 &0.954   &0.9513   &32.7    &33.26     \cr
			proposed             &\textbf{0.9566}  &\textbf{0.9595}   &\textbf{33.21}   &\textbf{34.05}    \cr
			\bottomrule
		\end{tabular}
		\label{ta2b}
	\end{threeparttable}
\end{table}

\subsection{Ablation Study of Different Methods}
The proposed method is compared with the model-based method and the data-driven one in Fig.  \ref{Data_Driven_Frame} as well as the method in \cite{1zheng2020}. The structure of the data-driven method is the same as the proposed method except the EAGB which requires the initial synthetic images. The average SSIM and PSNR for the $80$ sets of test images are shown in Table. \ref{ta2b}. The proposed method can improve the SSIM, PSNR, and the visual quality of the synthetic images. Clearly, the proposed method can make the synthetic images much closer to the ground truth images.  The results of MEF-SSIM \cite{MEFSSIM} are also shown in Table. \ref{tab2}. The quality of the finally fused images is also improved. The proposed EASRN outperforms the DNN in \cite{1zheng2020}. This also shows that the intrinsic properties of differently exposed images are important for the data-driven method on the saturation restoration.

\begin{figure}[htb]
	\centering
	\subfigure{
		\begin{minipage}[b]{0.465\linewidth}
			\includegraphics[width=1\linewidth]{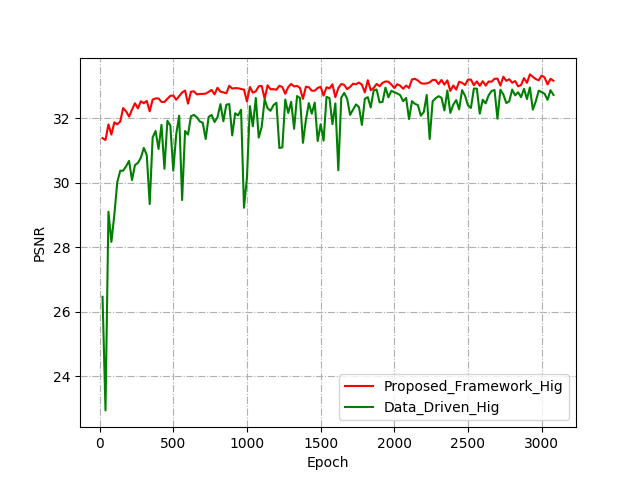}	
			\centerline{(a)}	
	\end{minipage}}
	\subfigure{
		\begin{minipage}[b]{0.465\linewidth}
			\includegraphics[width=1\linewidth]{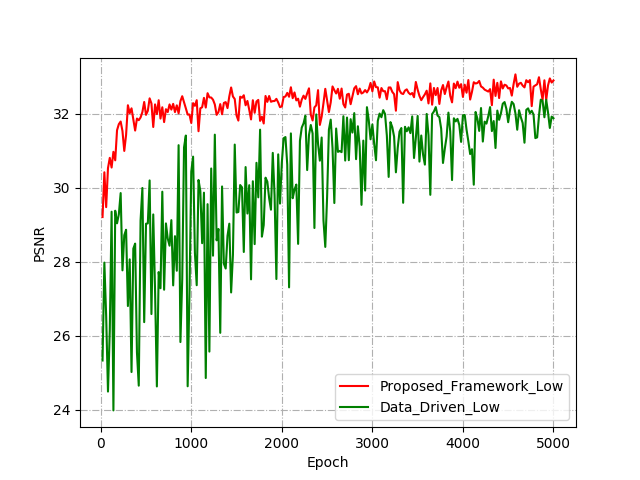}
			\centerline{(b)}
	\end{minipage}}
	\caption{(a) Comparisons of PSNR between the proposed framework and data-driven method for high exposure images. (b) Comparisons of PSNR between the proposed framework and data-driven method for low exposure images. The proposed framework is more stable than the data-driven method, and can also obtain higher  PSNR. }
	\label{Data_Driven}
\end{figure}

\begin{table}[htb]
	\begin{center}
		\centering
		\caption{ MEF-SSIM of Five Different Methods}
		\tabcolsep8pt\begin{tabular}{|c|c|c|c|c|c|}
			\hline
			image       & model             & Hybrid           & data        & proposed   & proposed   \\
			            &-based             & learning         &-driven      & (no EAGB)  &            \\\hline
			set1        & 0.9545            & 0.9610                  & 0.9604           &  0.9646         & \textbf{0.9661}\\
			set2        & 0.9419            & 0.9655                  & 0.9686           &  0.9717         & \textbf{0.9721}\\
			set3        & 0.9041            & 0.9488                  & 0.9489           &  0.9502         & \textbf{0.9519}\\
			set4        & 0.9476            & 0.9510                   & 0.9539           &  0.9561         & \textbf{0.9592}\\
			set5        & 0.9130            & 0.9166               & 0.9095           &  0.9171         & \textbf{0.9210}\\
			set6        & 0.9399            & 0.9408                    & 0.9394           &  \textbf{0.9422}         & 0.9414\\
			set7        & 0.9780            &\textbf{ 0.9795}                        & 0.9761           &  0.9765         & 0.9768\\
			set8        & 0.9243            & 0.9468      & 0.9489           &  \textbf{0.9450}         & 0.9433\\
			set9        & 0.8940            & 0.9034      & 0.9019           &  0.9047         & \textbf{0.9053}\\
			set10       & 0.9172            & 0.9447      & \textbf{0.9525}  &  0.9450         & 0.9361\\
			set11       & \textbf{0.9854}   & 0.9854      & 0.9852           &  0.9847         & 0.9852\\
			set12       & 0.8822            & 0.8944     & 0.8957           &  0.8993         & \textbf{0.9520}\\
			set13       & \textbf{0.9825}   & 0.9830     & 0.9809           &  0.9812         & 0.9816\\
			set14       & 0.9585            & 0.9672      & 0.9656           &  0.9661         & \textbf{0.9681}\\
			set15       & 0.9434            & 0.9432      & 0.9447           &  \textbf{0.9446}         & 0.9429\\
			set16       & 0.9828            & \textbf{0.9841}      & 0.9816           &  0.9820         & 0.9825\\
			set17       & 0.9176            & 0.9483      & \textbf{0.9492}  &  0.9407         & 0.9488\\
			set18       & 0.9052            & 0.9007      & 0.8859           &  0.8850         & \textbf{0.9065}\\
			set19       & 0.9020            & 0.9136     & 0.9185           &  0.9410         & \textbf{0.9332}\\
			set20       & 0.9430            & 0.9432     & 0.9442           &  0.9385         & \textbf{0.9452}\\\hline
			average     & 0.9362            & 0.9461      & 0.9460           &  0.9455         & \textbf{0.9510}\\\hline
		\end{tabular}
		\label{tab2}
	\end{center}
\end{table}

\begin{figure*}[htb]
	\centering
	\includegraphics[width=0.95\textwidth]{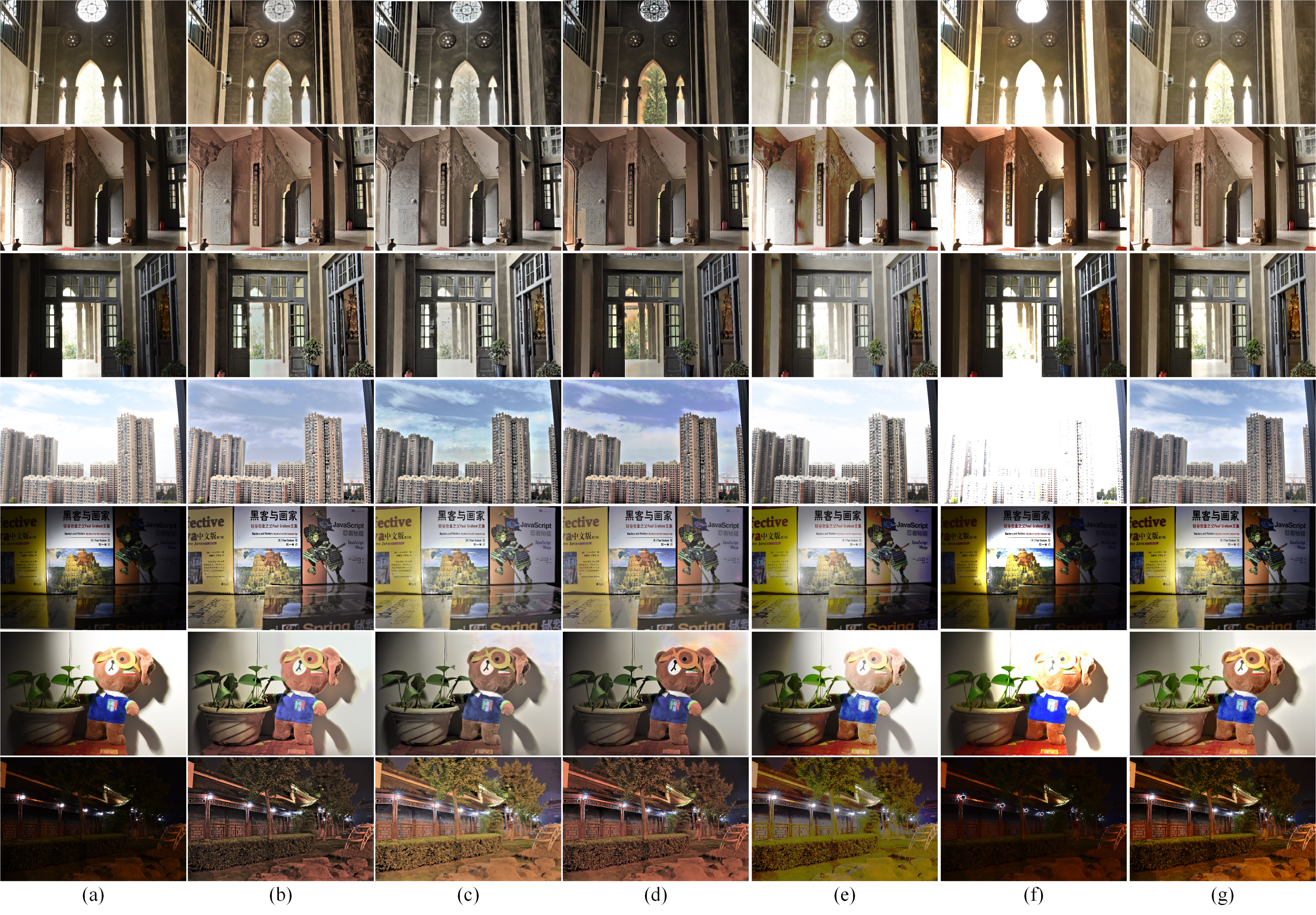}
	\caption{The first column shows the LDR images, the second  to  seventh columns illustrate the results of (b) ExpandNet \cite{ExpandNet}, (c) DrTMO \cite{DrTMO}, (d) Histogram HDR \cite{histogram}, (e) IRA HDR \cite{2015Pseudo}, (f) SingleHDR \cite{SingleHDR}, and (g) the proposed algorithm, respectively.}
	\label{Fig_seven}
\end{figure*}

The convergence speed is also analyzed. There is large difference between the input image $Z_1$ and the ground truth low/high exposure images $Z_{T_i}$. The low exposure images contain random noise, which is difficult to learn for the input image $Z_1$. On the other hand, the synthetic images $Z_i$ which are generated by the model-based method are already close to $Z_{T_i}$. It is easy for the residual CNNs to represent $(Z_{T_i} - Z_i)$, and the networks can converge much faster as shown in Fig. \ref{Data_Driven}. Clearly, the model-based method and the data-driven one indeed {\it compensate} each other in the proposed framework.

\subsection{Comparison of Six Different  Algorithms}

In this subsection,  the proposed saturation restoration algorithm is compared with three state-of-the-art (SOTA) data-driven algorithms including ExpandNet \cite{ExpandNet},  DrTMO \cite{DrTMO}, and SingleHDR \cite{SingleHDR}, as well as two SOTA mode-based algorithms including Histogram HDR \cite{histogram} and IRA HDR \cite{2015Pseudo}.  HDR images are generated by these three data-driven algorithms. Both the algorithms \cite{ExpandNet,DrTMO} are re-trained by using the dataset for the fair comparison. The training code of SingleHDR \cite{SingleHDR} is not available, thus the pre-trained model is used directly.  The proposed algorithm and the data-driven three algorithms are compared by using the mean squared error (MSE), the latest HDR-VDP-2 version 3.0, average PU2-SSIM and PU2-PSNR of HDR images \cite{HDR-VDP-2}.  The ground-truth HDR image is synthesized from the three differently exposed images $Z_{T_0}$, $Z_1$, and $Z_{T_2}$ by using the method in \cite{CRFs} and the CRFs $F_c(\cdot)$. The HDR image of the proposed algorithm is generated from $Z_0$, $Z_1$, $Z_2$, and the CRFs $F_c(\cdot)$. LDR images are produced by the two model-based algorithms. The proposed algorithm and the model-based algorithms are compared by using the MEF-SSIM with the reference images as $Z_{T_0}$, $Z_1$, and $Z_{T_2}$. It is shown in Tables \ref{tab3}, \ref{tab366}, \ref{tab367}, and \ref{tab368} that the the proposed algorithm usually outperforms the three data-driven algorithms from the MSE, HDR-VDP-2, PU2-SSIM, and PU2-PSNR points of view and the two model-based algorithms from the MEF-SSIM point of view.

\begin{table}[htb]
\begin{center}
\centering
\caption{MEF-SSIM of three Different Algorithms}
\tabcolsep8pt\begin{tabular}{|c|c|c|c|}\hline
			image &  Histogram HDR   &IRA HDR    & Proposed\\\hline
			set1  &  0.9246    & 0.8479         & \textbf{0.9661}\\
			set2  &  0.8855    & 0.5773         & \textbf{0.9721}\\
			set3  &  0.7945    & 0.5956         & \textbf{0.9519}\\
			set4  &  0.9021    & 0.8182         & \textbf{0.9592}\\
			set5  &  0.8695    & 0.7522         & \textbf{0.9210}\\
			set6  &  0.8921    & 0.7973         & \textbf{0.9414}\\
			set7  &  0.9300    & 0.8972         & \textbf{0.9768}\\
			set8  &  0.8710    & 0.7244         & \textbf{0.9433}\\
			set9  &  0.8712    & 0.8467         & \textbf{0.9053}\\
			set10 &  0.8473    & 0.7379         & \textbf{0.9361}\\
			set11 &  0.9338    & 0.9094         & \textbf{0.9852}\\
			set12 &  0.8609    & 0.8084         & \textbf{0.9520}\\
			set13 &  0.8730    & 0.7948         & \textbf{0.9816}\\
			set14 &  0.9401    & 0.8440         & \textbf{0.9681}\\
			set15 &  0.9305    & 0.9033         & \textbf{0.9429}\\
			set16 &  0.9545    & 0.7676         & \textbf{0.9825}\\
			set17 &  0.8917    & 0.8225         & \textbf{0.9488}\\
			set18 &  0.8549    & 0.5026         & \textbf{0.9065}\\
			set19 &  0.8668    & 0.8402         & \textbf{0.9332}\\
			set20 &  0.9119    & 0.7102         & \textbf{0.9452}\\\hline
		average   &  0.8903    & 0.7749         &\textbf{0.9510}\\\hline
\end{tabular}
\label{tab3}
\end{center}
\end{table}

\begin{table}[htb]
	\begin{center}
		\centering
		\caption{MSE of Four Different Algorithms}
		\tabcolsep8pt\begin{tabular}{|c|c|c|c|c|}\hline
			image & ExpandNet      & DrTMO           & SingleHDR          &  Proposed\\\hline
			set1  & 0.0033         & 0.0126          & 0.0243             & \textbf{5.9$\times${$10^{-4}$}}\\
			set2  & 0.0029         & 0.0043          & 0.0495             & \textbf{0.0014}\\
			set3  & 0.0096         & 0.0131          & 0.0481             & \textbf{0.0061}\\
			set4  & 0.0214         & 0.0133          & 0.0978             & \textbf{0.008}\\
			set5  & 0.0097         & 0.0359          & \textbf{3.6$\times${$10^{-4}$}}       & 7.2$\times${$10^{-4}$}\\
			set6  & 0.0134         & 0.0291          & 0.0149             & \textbf{0.0034}\\
			set7  & 0.0131         & 0.0528          & 0.0216             & \textbf{0.0019}\\
			set8  & 0.0228         & 0.0088          & 0.0443             & \textbf{0.012}\\
			set9  & 0.0126         & \textbf{0.0091} & 0.0098             & 0.0117\\
			set10 & \textbf{0.0232}& 0.0253          & 0.0673             & 0.024\\
			set11 & 0.0273         & 0.0419          & 0.0826             & \textbf{0.0116}\\
			set12 & 0.0027         & 0.0410          & 0.0081             & \textbf{0.001}\\
			set13 & 0.1162         & 0.2104          & \textbf{0.0097}             & 0.0351\\
			set14 & 0.0212         & 0.0462          & 0.0044             & \textbf{1.7$\times${$10^{-4}$}}\\
			set15 & 0.0518         & 0.0094          & 0.0057             & \textbf{0.0075}\\
			set16 & 0.0140         & 0.1044          & 0.0054             & \textbf{8.3$\times${$10^{-5}$}}\\
			set17 & 0.0074         & 0.0222          & 0.0580             & \textbf{0.0102}\\
			set18 & 0.0019         & 0.0011          & \textbf{7.2$\times${$10^{-4}$}}      & \textbf{1.1$\times${$10^{-4}$}}\\
			set19 & 0.0061         & 0.0162          & 0.0310             & \textbf{0.0057}\\
			set20 & 0.0347         & 0.0114          &\textbf{3.5$\times${$10^{-4}$}}                    &0.0026 \\\hline
		\end{tabular}
		\label{tab366}
	\end{center}
\end{table}

\begin{table}[htb]
	\begin{center}
		\centering
		\caption{HDR-VDP-2 of Four Different Algorithms}
		\tabcolsep8pt\begin{tabular}{|c|c|c|c|c|}\hline
			image & ExpandNet    & DrTMO     & SingleHDR              &  Proposed\\\hline
			set1  & 8.52         & 8.76      & 7.47          & \textbf{9.94}\\
			set2  & 8.45         & 8.38      & 8.39          & \textbf{9.32}\\
			set3  & 7.83         & 7.80      & 6.02            & \textbf{8.90}\\
			set4  & 7.11         & 7.92      & 5.78            & \textbf{8.85}\\
			set5  & 8.29         & 7.79      & 9.60            & \textbf{9.82}\\
			set6  & 7.86         & 7.14      & 7.50             & \textbf{9.42}\\
			set7  & 8.79         & 8.96      & 9.26           & \textbf{9.44}\\
			set8  & 7.84         & 8.37      & 7.52            & \textbf{8.81}\\
			set9  & 7.14         & 7.63      & \textbf{8.34}           & 8.30\\
			set10 & 7.11         & 7.46      & 6.47             & \textbf{8.30}\\
			set11 & 8.59         & 8.97      & 8.19           & \textbf{9.45}\\
			set12 & 8.68         & 7.22      & 7.16           & \textbf{9.60}\\
			set13 & 7.58         & 8.04      & 9.03            & \textbf{9.04}\\
			set14 & 7.97         & 8.24      & 9.01            & \textbf{9.85}\\
			set15 & 8.09         & 8.42      & 8.86            & \textbf{9.01}\\
			set16 & 8.08         & 7.72      & 7.82           & \textbf{9.88}\\
			set17 & 8.11         & 7.53      & 6.81           & \textbf{9.10}\\
			set18 & 8.37         & 8.15      & 7.74           & \textbf{9.85}\\
			set19 & 7.94         & 7.42      & 7.06           & \textbf{9.14}\\
			set20 & 7.14         & 8.93      & \textbf{9.65}           & 9.08\\\hline
		\end{tabular}
		\label{tab367}
	\end{center}
\end{table}

\begin{table}[htb]
	\centering
	\begin{threeparttable}
		\caption{Average PU2-SSIM and PU2-PSNR of Four different Methods.}
		\begin{tabular}{ccccccc}
			\toprule
			\multirow{2}{*}{}&
			\multicolumn{1}{c}{ PU2-SSIM }&\multicolumn{1}{c}{ PU2-PSNR }\cr
			\midrule
			ExpandNet   &0.8239     &23.08    \cr
			DrTMO       &0.7407    &24.51     \cr
			SingleHDR   &0.5150    &20.07     \cr
			Proposed    &\textbf{0.9596}    &\textbf{35.73}    \cr
			\bottomrule
		\end{tabular}
		\label{tab368}
	\end{threeparttable}
\end{table}

Fig. \ref{Fig_seven} includes the final images that are generated by the six algorithms. The pseudo-HDR images of the former three are tone mapped by using the state-of-the-art algorithm in \cite{1liang2018}. The algorithms in \cite{ExpandNet} and the proposed algorithm restore saturated regions better than those algorithms in  \cite{SingleHDR} \cite{DrTMO}\cite{histogram} \cite{2015Pseudo}. There are color distortions in the highlighted regions by the algorithms in \cite{ExpandNet}. As shown in the last row, there is brightness order reversal by the algorithm in \cite{2015Pseudo}. The images generated by the algorithms in \cite{histogram} \cite{DrTMO} and the proposed one look more natural and realistic.

\subsection{Limitation of the Proposed Algorithm}
\begin{figure}[!htb]
	\begin{center}
		\includegraphics[width=0.88\linewidth]{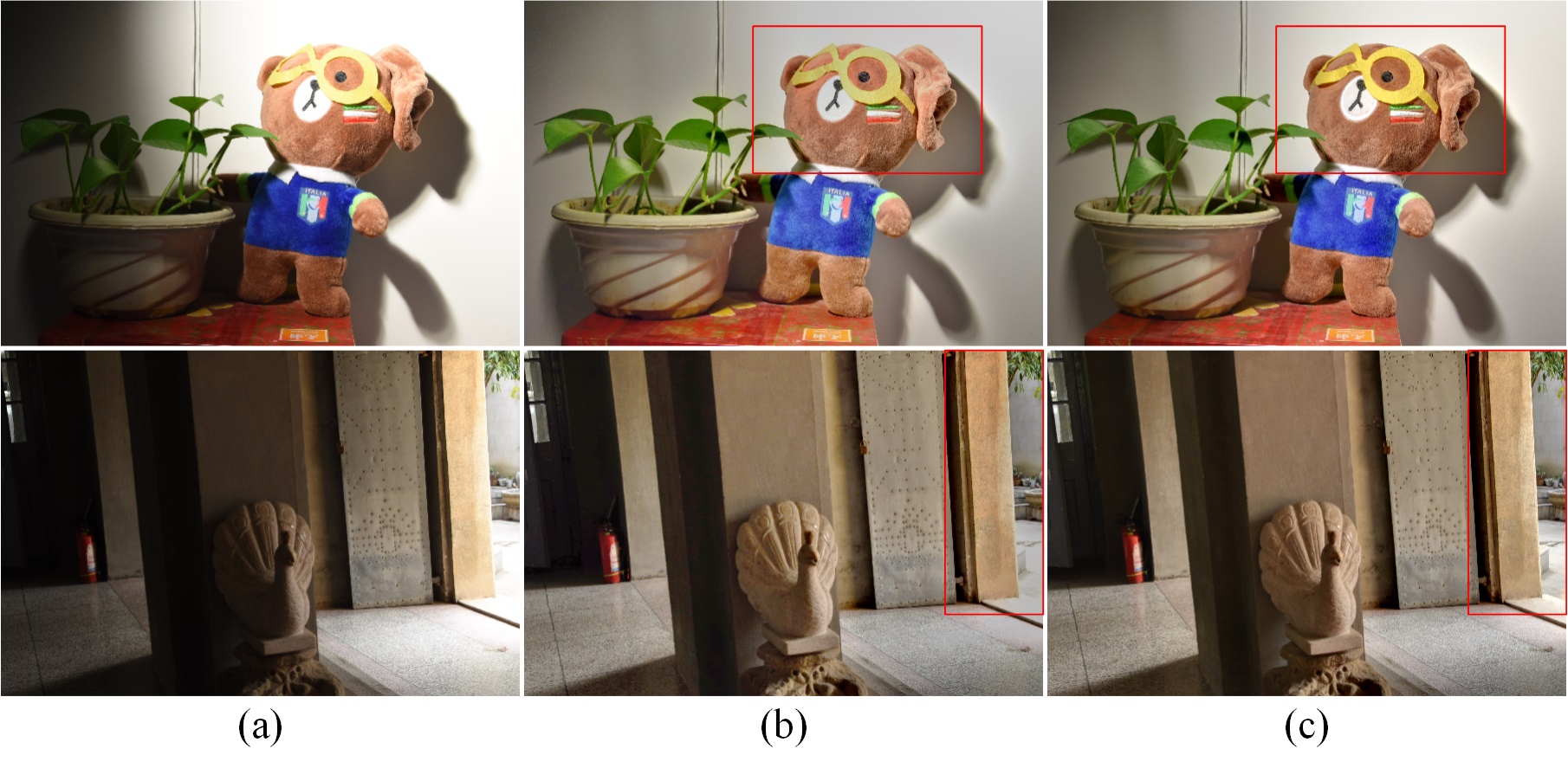}
	\end{center}
	\caption{Comparison of the information enriched LDR images by using inaccurate CRFs and accurate CRFs. (a) shows the input images; (b) illustrates the results by using inaccurate CRFs; (c) demonstrates the images by using accurate CRFs.}
	\label{Fig_nine}
\end{figure}

Although the proposed algorithm outperforms the other six algorithms, there is space for further study. For example, the proposed algorithm assumes  that the accurate CRFs are available. For images from unknown sources, it is quite challenging to estimate the  CRFs. If the estimated CRFs are not accurate,  the difference between the synthetic images and their ground truth images become large, and the quality of the brightened images drops a little bit, as shown in Fig. \ref{Fig_nine}.

\section{Conclusion Remarks and Discussions}
\label{paradigm5}
A new learning framework is introduced to restore saturated regions for a low dynamic range (LDR) image of a high dynamic range (HDR) scene. Two initial synthetic images are first generated by using the model-based method and they are then enhanced by using the data-driven residual convolutional neural networks. They {\it compensate} each other very well in such a neural augmentation.  The proposed framework has a potential to achieve learning with few data. All the input image and the two synthetic images are finally combined together to produce an LDR or HDR image.

The on-line cost of the proposed neural augmentation is slighter higher than that of the pure deep learning method due to the inclusion of the model-based method. On the other hand,  a complexity scalable brightening algorithm is provided by the proposed framework. An image is produced instantly by the simple model-based method for previewing on the mobile device, and a higher quality image is then produced to replace the previewed one after a while. The proposed algorithm is also applicable to study saturation restoration of LDR videos.  Flicking artifacts could be well addressed because of the model-based method. One more interesting application of the proposed algorithm is image stitching. The intensity mapping functions (IMFs) can be estimated using the overlapping areas. Instead of only correcting color as in \cite{1ding2021}, a HDR panorama image will be produced. These problems will be studied in our future research.

\end{document}